\documentclass[10pt,twocolumn,letterpaper]{article}

\usepackage{cvpr}              % To produce the CAMERA-READY version
\usepackage{graphicx}
\usepackage{amsmath}
\usepackage{amssymb}
\usepackage{booktabs}
\usepackage{multirow}
\usepackage{adjustbox}

\newcommand{\floor}[1]{\lfloor #1 \rfloor}
\usepackage[pagebackref,breaklinks,colorlinks]{hyperref}

\usepackage[capitalize]{cleveref}
\crefname{section}{Sec.}{Secs.}
\Crefname{section}{Section}{Sections}
\Crefname{table}{Table}{Tables}
\crefname{table}{Tab.}{Tabs.}

\def\cvprPaperID{8200} % *** Enter the CVPR Paper ID here
\def\confName{CVPR}
\def\confYear{2022}

\begin{document}
	
	\title{Voxel Set Transformer: A Set-to-Set Approach to \\ 3D Object Detection from Point Clouds}
	
	\author{
		Chenhang He,  Ruihuang Li,  Shuai Li, Lei Zhang\thanks{Corresponding author.} \\
		The Hong Kong Polytechnic University\\
		{\tt\small \{csche, csrhli, csshuaili, cshzeng\}@comp.polyu.edu.hk}}
	
	\maketitle
	
	%%%%%%%%% ABSTRACT
	\begin{abstract}
		Transformer has demonstrated promising performance in many 2D vision tasks. However, it is cumbersome to compute the self-attention on large-scale point cloud data because point cloud is a long sequence and unevenly distributed in 3D space. To solve this issue, existing methods usually compute self-attention locally by grouping the points into clusters of the same size, or perform convolutional self-attention on a discretized representation. However, the former results in stochastic point dropout, while the latter typically has narrow attention fields. In this paper, we propose a novel voxel-based architecture, namely Voxel Set Transformer (VoxSeT), to detect 3D objects from point clouds by means of set-to-set translation. VoxSeT is built upon a voxel-based set attention (VSA) module, which reduces the self-attention in each voxel by two cross-attentions and models features in a hidden space induced by a group of latent codes. With the VSA module, VoxSeT can manage voxelized point clusters with arbitrary size in a wide range, and process them in parallel with linear complexity. The proposed VoxSeT integrates the high performance of transformer with the efficiency of voxel-based model, which can be used as a good alternative to the convolutional and point-based backbones. VoxSeT reports competitive results on the KITTI and Waymo detection benchmarks. The source codes can be found at \url{https://github.com/skyhehe123/VoxSeT}. 
	\end{abstract}
	\
	%%%%%%%%% BODY TEXT
	\section{Introduction}
	\label{sec:intro}
	
	Object detection from 3D point cloud has been receiving extensive attention as it empowers many applications like autonomous driving, robotics and virtual reality. Unlike 2D images, 3D point clouds are naturally sparse and unevenly distributed in continuous space, impeding the CNN layers from being directly applied. To resolve this issue, some approaches \cite{second, voxelnet, sa-ssd, cia-ssd, voxel-rcnn} first transform the point cloud into a discrete representation and then apply CNN models to extract high dimensional features. Another class of approaches \cite{pointrcnn, std, 3dssd, MLCVNet, frustum} model the point cloud in continuous space, where the multi-scale features are extracted through interleaved grouping and aggregation steps. 
	
	\begin{figure}
		\centering
		\includegraphics[width=0.98\columnwidth]{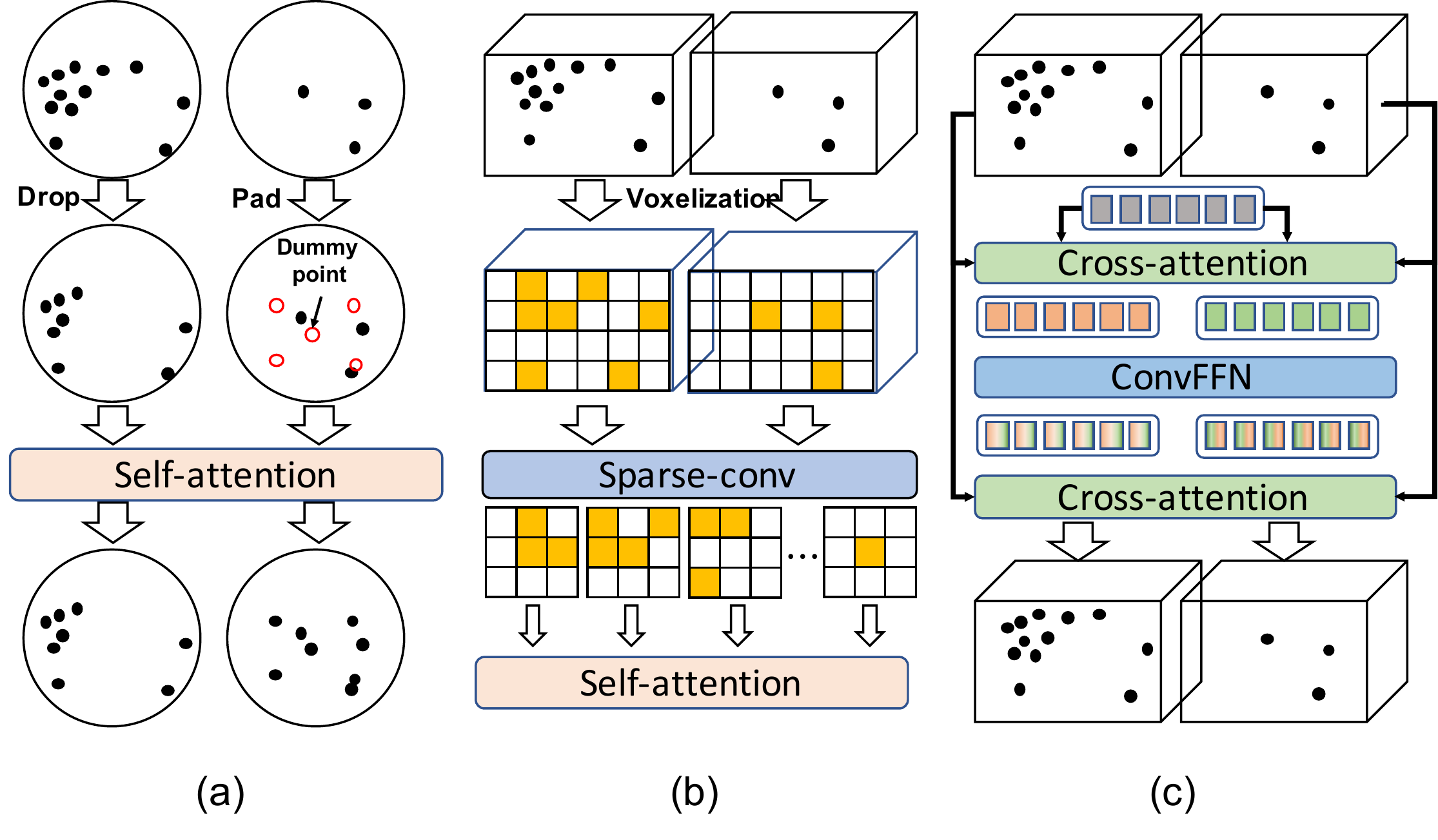}
		\vspace{-3mm}
		\caption{The illustrations of (a) grouping-based, (b) convolutional-based and (c) our proposed induced set-based attention mechanisms.}
		\vspace{-5mm}
		\label{fig:motivate}
	\end{figure}

	Beyond the above two schemes, transformer-based models \cite{pct, pointformer, pointTr, votr, ct3d, 3detr} have recently attracted great interest in processing point cloud data as the self-attention used in transformers is invariant to permutation and cardinality of the input components, which makes transformer an appropriate choice for point cloud processing.
	The main limitation of transformer models, however, lies in that the self-attention computation is quadratic. Each token has to be updated by using all the other tokens from previous layers, making self-attention intractable for long sequence point clouds. Point Transformer \cite{pointTr} builds transformers upon a PointNet\cite{pointnet} architecture, which hierarchically groups the point cloud data into different clusters and computes self-attention in each cluster. CT3D \cite{ct3d} presents a two-stage point cloud detector, where 3D RoIs are extracted to group the raw points in the first stage and transformers are applied to the grouped points in the second stage. 
	
	However, since the distribution of point clouds is extremely uneven, the number of points in each cluster varies a lot. To enable the self-attention to run in parallel, current approaches \cite{ct3d, pointTr,pointformer} balance the token number in each cluster by stochastically dropping points or padding dummy points (see Figure \ref{fig:motivate}(a)). This results in unstable detection results and redundant computations. Besides, each operation of grouping $n$ points to $m$ clusters will cost $\mathcal O(nm)$ complexity, which is relatively intensive. Alternatively, Voxel Transformer \cite{votr} performs self-attention on a discrete voxel grid, as depicted in Figure \ref{fig:motivate}(b). It computes self-attention in a convolutional manner and hence is as efficient as sparse convolution with $\mathcal O(n)$ complexity. However, since convolutional attention is a point-wise operation, to save the memory, the attention field of the convolutional kernel is typically small, thus hindering the voxel transformer to model long-range dependencies. It is worth mentioning that though Group-free \cite{group-free} and 3DETR \cite{3detr} present a promising solution by computing self-attention on a reduced set of seed points, this solution is only applicable to indoor scenes, where the point clouds are relatively dense and concentrated. Considering that the point clouds of outdoor scenes are typically sparse, large-scale (\eg, $>20k$), and unevenly distributed, the scale and coverage of seed points remain an issue.
	
	To address the above issues, we introduce a voxel-based set attention (VSA) module. For each VSA, we divide the whole scene into non-overlapping 3D voxels and compute the voxel indices of the input point with instant efficiency. We use these voxels to determine the attentive region which is analogous to the window attention in
	SwinTransformer \cite{swin}. Unlike image, LiDAR has irregular structures, and the resulting attention groups have different lengths, which hinders the parallelization of the model.

	Inspired by the induced set transformer \cite{setTr}, we assign a group of trainable ``latent codes'' to each voxel. These latent codes build a fixed-length bottleneck for the point cloud, through which the information from input points within the voxel can be compressed to a static hidden space. This formulation is based on the key observation that the self-attention matrix is typically low-rank, and hence we can decompose an intensive full self-attention into two consecutive cross-attention modules. As shown in Figure \ref{fig:motivate}(c), VSA first transforms the latent codes, which serve as queries, to a hidden space by attending to the projected features, \ie, keys and values, from the input points. The transformed hidden features, which encode the context information of the input points in each voxel, are enriched by a convolutional feed-forward network, in which the features across voxels exchange their information in spatial domain. After that, the hidden features are attentively fused with input, producing output features of the input resolution. By leveraging the latent codes, the cross-attention performed in all voxels can be vectorized, making VSA a highly parallel module. Given $n$ $d$-dimensional input features and $k$ latent codes, VSA has a complexity of $\mathcal O(nkd)$ and it can be implemented with general matrix multiplications.
	
	With VSA, we propose a Voxel Set Transformer (VoxSeT) to detect 3D objects by learning point cloud features in a set-to-set translation process. VoxSeT is composed of VSA modules, MLP layers and a shallow CNN for Birds-Eye-View (BEV) feature extraction. To verify the effectiveness of the proposed model, we conduct experiments on two 3D detection benchmarks, KITTI and Waymo open dataset. VoxSeT achieves competitive performance with current state-of-the-arts. In addition, the proposed VSA module can be seamlessly adopted into point-based detectors such as PointRCNN \cite{pointrcnn}, and demonstrates advantages over the set abstraction module.
	
	In summary, in this work we first invent a voxel-based set attention module, which can model long-range dependencies from the token cluster of arbitrary size, bypassing the limitation of current grouped-based and convolution-based attention modules. We then present a Voxel Set Transformer to learn point cloud features effectively by leveraging the superiority of transformer on large-scale sequential data. Our work provides a novel alternative to the current convolutional and point-based backbones for 3D point cloud data processing. 
	
	\section{Related work}
	\subsection{3D object detection from point clouds}
	Early approaches on 3D object detection from point cloud can be categorized into two classes. The first class of methods transform the point cloud into more compact representations, \eg, Birds-Eye-View (BEV) images \cite{complex,mv3d,avod}, frontal-view range images \cite{rangedet, rangeioudet, to-the-point}, and volumetric features \cite{fcn3d,pixor, voxelnet}.  Yan \textit{et al.} \cite{second} developed a sparse convolutional backbone to efficiently process the point clouds by encoding the point clouds into a 3D sparse tensor. Lang \textit{et al.} \cite{pointpillars} further accelerated the detection rate by stacking the voxel features as a ``pillar'' and using 2D CNN to process. Another class of methods \cite{pointrcnn, votenet, frustum, std, 3dssd} process the point cloud in a continuous space by employing a PointNet\cite{pointnet++} architecture. The point-wise features in multi-scales are extracted in stages with interleaved grouping and sampling operations. Shi \textit{et al.} \cite{pointrcnn} and Yang \textit{et al.} \cite{std} proposed to generate 3D RoIs from PointNet outputs and apply the RoIs to group point-wise features for further refinement. Qi \textit{et al.} \cite{houghvoting} proposed a deep voting method to cluster the points from objects' surface to detect the object with insufficient points. Unlike compact representations, point-wise features preserve more details and fine-grained structures of original point clouds. Based on this fact, some approaches \cite{pvrcnn, sa-ssd} employ a hybrid representation in both point and voxel spaces to achieve more reliable detection outputs. 
	
	Our proposed architecture is largely motivated by voxel-based approaches. We partition the point cloud into voxel grid and execute self-attention locally, endowing our model with inductive bias and computational efficiency.
	
	\subsection{Transformer in point cloud analysis}
	Recently, Transformer \cite{transformer} has demonstrated its great success in many computer vision tasks such as image classification \cite{vit, deit}, 2D object detection \cite{detr, swin}, and other dense prediction tasks \cite{dpt, segTr}. For point cloud analysis, Zhao \textit{et al.} \cite{pointTr} proposed a novel subtraction attention based operator for point cloud classification and segmentation. Guo \textit{et al.} \cite{pct} investigated a dual attention to process the point clouds in feature and edge space. Misra \textit{et al.} \cite{3detr} and Liu \textit{et al.} \cite{group-free} used transformer to process point clouds as sequential data, preventing the models from stacking hierarchical grouping and sampling modules. Miao \textit{et al.} \cite{votr} embeded self-attention into a sparse convolutional kernel. Sheng \textit{et al.} \cite{ct3d} built the transformer on top of a two-stage detector and operated attention on the points grouped by RoIs. 
	
	Unlike the above approaches that perform self-attention on a fixed-size token cluster, our proposed Voxel Set Transformer leverages the idea of induced set transformer \cite{setTr} to decompose self-attention into two cross-attentions, making it possible to perform self-attention on the token clusters of arbitrary size.
	
	\begin{figure*}
		\centering
		\includegraphics[width=0.98\linewidth, height=5.5cm]{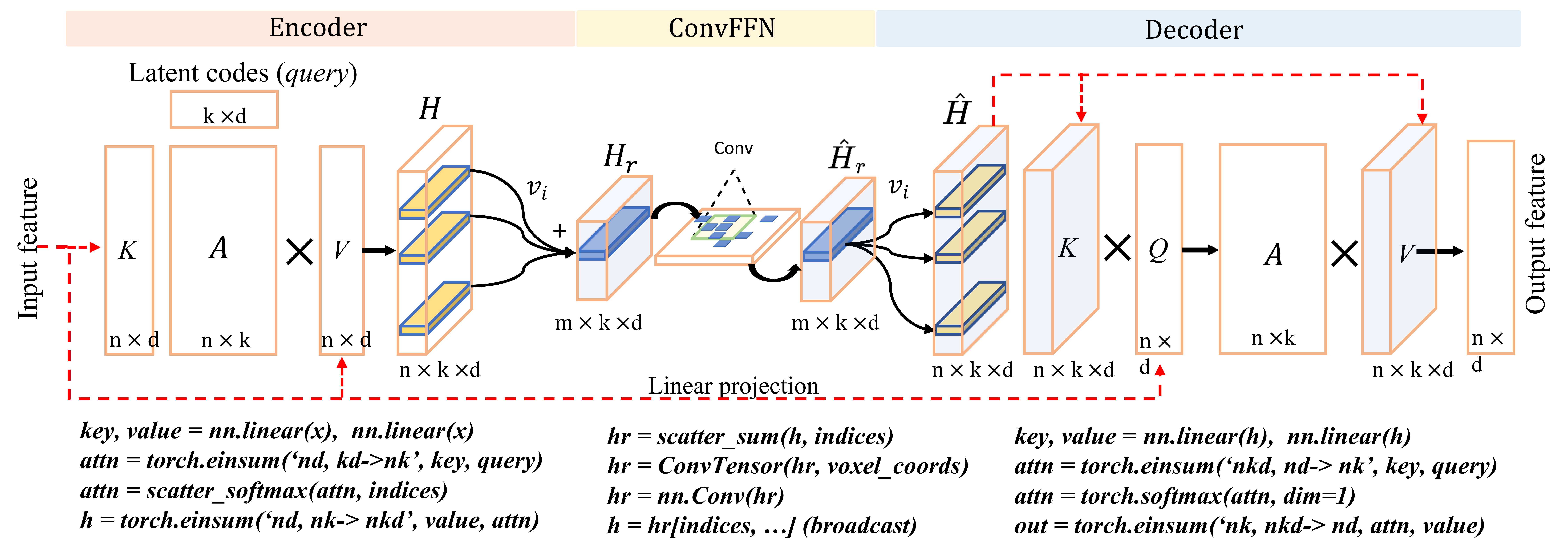}
		\vspace{-2mm}
		\caption{The matrix multiplication used in implementing the voxel-based set attention (VSA) module. The pseudo codes in PyTorch-stype of each step are presented bellow the diagram. }
		\label{fig:matrix}
		\vspace{-3mm}
	\end{figure*}

	\section{Methodology}
	
	%We briefly review the prior induced set transformer models \cite{setTr}, they conceptual similarity to voxel-based set attention module (VSA) and how it can be efficiently executed in parallel with matrix multiplication. Then, we introduce the overall architecture of Voxel Set Transformer (VoxSeT) model and the detection objectives.
	
	\subsection{Preliminary}
	
	It is prohibitive to directly apply self-attention on point cloud data due to its quadratic computational complexity. To bypass the issue, an induced set attention block was proposed in \cite{setTr}, where the full self-attention in a set was approximated by two reduced cross-attentions induced by a group of latent codes. Given an input set $X\in \mathcal{R}^{n\times d}$ of size $n$ with dimension $d$ and $k$ latent codes $L \in \mathcal{R}^{k \times d}$, the output set $O\in \mathcal{R}^{n\times d}$ from the induced set attention block can be formulated as 
	\begin{align}
		H & = \text{CrossAttention}(L, X) \in \mathcal{R}^{k\times d}, \label{eq:crossAttn1} \\
		\hat H & = \text{FFN}(H) \in \mathcal{R}^{k\times d},\\
		O & = \text{CrossAttention}(X, \hat H) \in \mathcal{R}^{n\times d}.	\label{eq:crossAttn2}
	\end{align}
	The first cross attention transforms the latent features $L$ into hidden features $H$ by attending to the input set. This step costs $\mathcal O(nkd)$ complexity, which is linear to $n$ as the number of latent codes $k$ is fixed and usually very small. The transformed hidden features contain information about the input set $X$ and then they are updated by a point-wise feed-forward network (FFN).  This point-wise operation costs $\mathcal O(k)$ complexity and it learns highly semantic features from the input set. The second cross attention attends the input set to the resulting hidden features, which costs $\mathcal O(nkd)$ complexity, producing an output set of length $n$. The induced set attention is based on the assumption that the self-attention can be approximated with low-rank projections, thus the self-attention can be regarded as performing a $k$-clustering on the inputs where the latent codes serve as cluster centers. This is also analogous to the clustered attention \cite{cluster-attn} and \textit{Linformer} \cite{linformer}, where the input set is explicitly reduced with linear projection. 
	
	\subsection{Voxel-based Set Attention (VSA)}
	\label{sec:vsa}
	Unlike images, point clouds are widely distributed and have weak semantic associations in scene level, while they have strong structural details in the local region.
	Instead of compressing all the input points into a hidden space, we modify the above induced set attention to be performed locally. Specifically, we partition the scene into a voxel grid and assign a set of latent codes to each voxel. We refer to the module as \textit{Voxel-based Set Attention} (VSA).
	
	%\noindent 
	\textbf{Scatter kernel function.}
	As mentioned before, VSA is a highly parallel module, where the operations across voxels can be vectorized. This vectorization can be achieved by the \textit{scatter} function\footnote{https://github.com/rusty1s/pytorch\_scatter}, which is a cuda kernel library that performs symmetric reduction, \eg, sum, max and mean, on different segments of a matrix. In our case, we regard the input set as a single matrix, each row of which corresponds to a point-wise feature, and its belonging voxel can be indexed by a table of voxel coordinates.
	
	Let $\{p_i=(x_i, y_i, z_i):i=1,...n\}$ denote the coordinates of point cloud and $[d_x, d_y, d_z]$ be the voxel size in three dimensions. The voxel coordinates $\mathcal V$ can be computed by
	$\mathcal V=\{
	\mathcal V_i =(\floor{\frac{x_i}{d_x}}, \floor{\frac{y_i}{d_y}}, \floor{\frac{z_i}{d_z}}):i=1,...,n
	\}
	$,
	where $\floor \cdot$ is the floor function. Hence, given point-wise input features $\{X_i:1=1,...n\}$, their reduced voxel-wise form $\{Y_{j}:j=1,...,m\}$  after a symmetric function $\text{F}(\cdot)$ can be represented as:
	\begin{equation}
		Y = \{ \text{F} (\{X_i: \mathcal V_i=j\}):j=1, ..., m\}
	\end{equation}
	where $m$ is the number of non-empty voxels. With $scatter$ function $\text{F}_{scatter}$, the above equation can be written in a vectorized form, \ie,
	\begin{equation}
		Y= \text{F}_{scatter}(X, \mathcal V).
	\end{equation}
	
	\noindent By deploying VSA, we do not need to stochastically drop or pad the points in each voxel and the complexity of the model is linear. 
	
	In Figure \ref{fig:matrix}, we illustrate the VSA in a matrix-multiplication form for ease of comprehension. As can be seen, the module is analogous to an encoder-decoder architecture, where the input set is encoded to a hidden space, then the hidden features are refined through a ConvFFN and finally decoded to produce the output set.

	%\noindent 
	\textbf{Encoder.} In the encoder, we first project the input features from the previous module to the key $K\in \mathcal{R}^{n\times d}$ and the value $V\in \mathcal{R}^{n\times d}$ with linear projections, respectively. Next, we perform cross attention between the key and the latent codes (query) $L \in \mathcal{R}^{k \times d}$, producing the attention matrix $\mathcal A\in \mathcal{R}^{n\times k \times d}$. The attention matrix $\mathcal A$ is then normalized voxel-wisely to obtain $\tilde{ \mathcal A}$, and multiplied with the value, producing hidden features $\tilde H$. The calculation of $\tilde H$ can be formulated as:
	\begin{align}
		\tilde{\mathcal A} &= \text{Softmax}_{scatter}( \mathcal A, \mathcal V)  ,\;  \mathcal A = KL^T, \\
	    H&= \tilde{\mathcal A} ^TV.
	\end{align}
	After that, we perform voxel-wise reduction on the hidden features based on the voxel indices $\mathcal V$:
	\begin{equation}
		H_r =  \text{Sum}_{scatter}(H, \mathcal V).
	\end{equation}
	The overall computations in the encoder include two GEMMs and two $scatter$ operations. The overall complexity of the encoder is $\mathcal{O}(2n(k+1)d)$.

	It is worth mentioning that the cross-attention based encoding scheme can be viewed as an extension of voxel feature encoding (VFE) used in \cite{voxelnet, second, pointpillars}. The difference is that VFE encodes the points within a voxel into a single feature vector, while our scheme encodes the points based on a \textit{codebook} consisting of the latent features. Owing to the high expressive power of VSA, we can use a relatively large voxel size to capture the features in a wide range.

	\begin{figure*}
		\centering
		\includegraphics[width=0.99\linewidth]{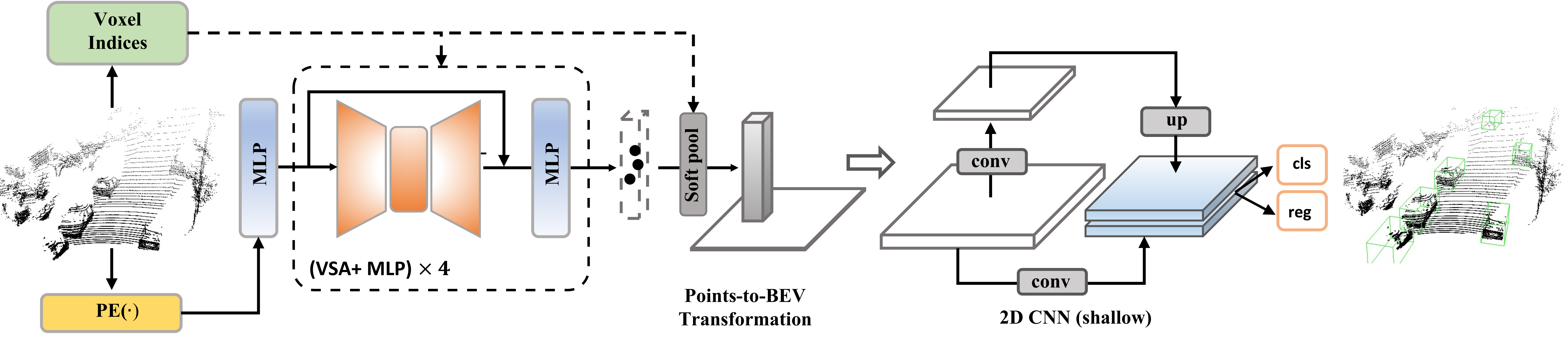}
		\vspace{-2mm}
		\caption{The overall architecture of the proposed Voxel Set Transformer (VoxSeT).}
		\vspace{-3mm}
		\label{fig:model}
	\end{figure*}

	%\noindent 
	\textbf{Convolutional feed-forward network.} The core idea of VSA is to encode the region-wise features into a hidden space using latent codes. The hidden features work as a bottleneck, through which we apply a ConvFFN to achieve more flexible and complex information update. Unlike conventional FFN that only performs point-wise token update, ConvFFN enables the information exchange accross voxels, which is especially important for dense prediction. To adaptively integrate voxel features with global dependencies, we scatter the reduced hidden features into a 3D sparse tensor based on their voxel coordinates $C_r$, and then conduct two depth-wise convolutions (DwConv) on them to enforce the feature interaction in spatial domain. Given convolutional weights $W_1$ and $W_2$, the enriched hidden features $\hat H_r \in \mathcal R^{m\times k \times d}$ from ConvFFN can be written as:
	\begin{equation}
		\hat H_r = \text{DwConv}(\sigma(\text{DwConv}(\mathcal T(H_r, C_r); W_1) );W_2), 
	\end{equation}
	where $\sigma$ denotes the non-linear activation, $\mathcal T$ denotes the formulation of sparse tensor, and the number of groups in DwConv equals to that of latent features. This operation costs $\mathcal{O}(\frac{HWDkd}{d_x d_y d_z})$ complexity, where $H, W, D$ refer to the point cloud range in three directions, respectively, and $[d_x, d_y, d_z]$ specifies the voxel size. The ConvFFN plays an important role in VSA as it introduces desirable inductive bias and global context to the module. More studies on ConvFNN will be discussed in Sec \ref{sec:ablation}.

	%\noindent 
	\textbf{Decoder.} The decoder reconstructs the output set from the enriched hidden features $\hat H_r$. Specifically, we first broadcast the hidden features based on the voxel indices $\mathcal V$, producing $\hat H \in \mathcal R^{n\times k \times d}$ which has the same length as the input set. Then we generate the query, key-value pair from the input set and the hidden features, respectively, with linear projections. Given matrices of query $Q \in \mathcal{R}^{n\times d}$, key $K \in \mathcal{R}^{n\times k \times d}$ and value $V \in \mathcal{R}^{n\times k \times d}$, the decoder output $O$ can be calculated as:
	\begin{align}
		\mathcal A &= [ \mathcal A_1, ..., \mathcal A_n] = [K_1Q_1^T, ..., K_nQ_n^T ],\\
		\tilde{\mathcal A} &= [\text{Softmax}(\mathcal A_1), ..., \text{Softmax}(\mathcal A_n)],\\
		O &= [O_1, ..., O_n] = [\tilde{\mathcal A}_1^TV_1, ..., \tilde{\mathcal A}_n^TV_n] 
	\end{align}

	The overall computational complexity of the decoder is $\mathcal{O}(2nkd)$. Owing to the flexibility of cross-attention mechanism, VSA formulates the point cloud processing as a set-to-set translation problem.

	%\noindent 
	\textbf{Relative position embedding}. As discussed in \cite{pointnet++, pointformer}, preserving the local structure of point cloud is crucial to improve performance. Therefore, we introduce a Positional Embedding (PE) module to encode the local coordinates of the point clouds within a voxel to a high dimensional feature and inject them into each VSA module. Specifically, the PE module applies the Fourier parameterization to take values $[sin(f_k \pi x), cos(f_k \pi, x)]$, given the normalized local coordinates $x \in [0, 1]$ and the $k^{th}$ frequency $f_k$ with bandwidth $L$. The resulted Fourier embedding, which has a dimension of $3L$, is further mapped to the input dimension of the first MLP module through a learnable linear layer.

	\subsection{Voxel Set Transformer (VoxSeT)}
	\label{sec:voxset}
	The overall architecture of VoxSeT is illustrated in Figure \ref{fig:model}.
	Following the traditional transformer paradigm, the VoxSeT backbone is composed of inter-connected multi-layer perception (MLP) and VSA modules. We use batch norm as the normalization layer and wrap each VSA module into a residual block for optimal gradient flow.
	
	Unlike grouping-based approaches \cite{pointnet++, votenet} that progressively downsample and aggregate point-wise features for context extraction, our backbone extracts point cloud features as a set-to-set translation process. The semantic level of the features is controlled by the size of voxels in the VSA module. We empirically found that applying large voxels can learn richer context information, and demonstrate better understanding of the objects with sparse points, especially pedestrian and cyclist instances. We will present the settings of VSA modules in Sec. \ref{sec:settings}. 
	
	%\noindent 
	\textbf{Birds-eye-view feature encoding.} In point cloud detection, a common phenomenon is that models using dense birds-eye-view (BEV) features \cite{second, voxel-rcnn, fast-pointrcnn} generally achieve higher recall than those using sparse point-wise features \cite{pointrcnn, std}. In this regard, we encode the point-wise features from the backbone into a BEV representation and apply a shallow CNN to increase the feature density. The CNN has only two strides, each involving three convolutions. The convolutional features of two strides are finally concatenated and passed to the detection head for bounding-box prediction. To generate BEV features, we aggregate the point-wise features within a pillar of size 0.36m$\times$0.36m and apply a ``soft-pooling'' operation to produce features in BEV. Given point-wise output features $X^j\in \mathcal R^{k\times d}$ in the $j^{th}$ pillar, the pillar-wise features after pooling $F^j$  can be formulated as:
	\begin{align}
		F^j &= \sum_{m=1}^kX^j_m * w^j_m, \;  \;    w^j_m = \frac{e^{X^j_m}}{\sum_{m=1}^ke^{X^j_m}}.
	\end{align}

	%\noindent 
	\textbf{Detection head and training objectives.} To enhance the expressiveness of VoxSeT backbone, we follow PointRCNN \cite{pointrcnn} to apply the foreground segmentation loss $\mathcal L_{seg} $ to the output features. This forces VoxSeT to capture contextual information for generating accurate bounding-boxes. The detection head follows the traditional anchor-based design \cite{second,pointpillars}. The final loss then becomes:
	\begin{equation}
		\mathcal L = \mathcal L_{seg} +\frac{1}{N_p}( \mathcal L_{cls} + \mathcal L_{reg}) + \mathcal L_{dir} ),
	\end{equation}
	where $N_p$ is the number of positive samples whose IoU with anchors lies between $[\sigma_1, \sigma_2]$. $\mathcal L_{\text{cls}}$ is the focal loss for bounding-box classification and $\mathcal L_{\text{reg}}$ is the Smooth-$L_l$ loss for bounding-box offsets regression. $\mathcal L_{dir}$ is a binary entropy loss for bounding-box orientation prediction. The readers are referred to \cite{pointrcnn} and \cite{second} for details.

	\textbf{Two-stage model.}
	It is worth noting that VoxSeT can be extended to a two-stage detector, in which we employ the efficient RoI head from LiDAR-RCNN \cite{lidar-rcnn} as our second-stage module. To more clearly illustrate our contribution, we also report the performance of a single-stage detector, and our VoxSeT demonstrates superior performance over the current single-stage baselines.

	\begin{table*}[t]
		\caption{Performance comparison with state-of-the-art methods on the Waymo dataset with 202 validation sequences ($\sim40k$ samples) for vehicle detection.}
		\vspace{-2mm}
		\centering
		
		\begin{adjustbox}{width=0.96\textwidth}
			\begin{tabular}{c|c||c|ccc||c|ccc}
				\hline
				\multirow{2}{*}{Method} &
				\multirow{2}{*}{Backbone} &
				\multicolumn{4}{c||}{3D mAP }  &
				\multicolumn{4}{c}{BEV mAP }  \\
				
				\cline{3-10}
				
				{}  &{} & Overall & 0-30m  & 30-50m & 50m-inf  & Overall & 0-30m  & 30-50m & 50m-inf  \\ \hline\hline
				\textit{\textbf{LEVEL\_1 (IoU=0.7):}}& & & &&&&&& \\
				PointPillar \cite{pointpillars} (CVPR19) &CNN &56.62& 81.01 &51.75 &27.94 &75.57 &92.10& 74.06 &55.47  \\
				MVF \cite{dy-voxelnet} (CoRL20) &CNN & 62.93 &86.30 &60.02 &36.02 &80.40& 93.59& 79.21 &63.09  \\
				PV-RCNN \cite{pvrcnn} (CVPR20) &SpCNN & 70.30 &91.92 &69.21 &42.17 &82.96 &97.35& 82.99 &64.97  \\
				Voxel-RCNN \cite{voxel-rcnn} (AAAI21)  &SpCNN & 75.59& 92.49 &74.09 &53.15 &88.19 &97.62& 87.34 &77.70  \\
				VoTR-TSD \cite{votr} (ICCV21)  &Transformer& 74.95 & 92.28 & 73.36 & 51.09 & - & - & - & - \\
				CT3D \cite{ct3d} (ICCV21)  &SpCNN &76.30 &92.51 &75.07 &\textbf{55.36} & \textbf{90.50} &\textbf{97.64} &88.06& \textbf{78.89} \\

				\textit{VoxSeT (ours)} &Transformer & 76.02 &91.13 &75.75&54.23 &89.12 &95.12&87.36& 77.78 \\
				\textit{VoxSeT + CT3D (RoI head)}  &Transformer& \textbf{77.82} &\textbf{92.78} &\textbf{77.21}&54.41 &90.31 &96.11&\textbf{88.12}& 77.98
				
				\\ \hline \hline
				
				\textit{\textbf{LEVEL\_2 (IoU=0.7):}} && & &&&&&& \\
				
				PV-RCNN \cite{pvrcnn} (CVPR20)  &SpCNN &  65.36 &91.58 &65.13 &36.46 &77.45 &94.64& 80.39 &55.39 \\
				Voxel-RCNN \cite{voxel-rcnn} (AAAI21)  &SpCNN & 66.59 &91.74 &67.89 &40.80 &81.07 &96.99& 81.37& 63.26 \\
				VoTR-TSD \cite{votr} (ICCV21) &Transformer &65.91 &-&- &- &-& - &-&- \\
				CT3D \cite{ct3d} (ICCV21) &SpCNN &69.04 &91.76&68.93 &42.60 &\textbf{81.74}& \textbf{97.05} &\textbf{82.22}& \textbf{64.34} \\

				\textit{VoxSeT (ours)} &Transformer& 68.16&91.03 &67.13&42.23 &76.13 &94.13&81.78& 58.13 \\
				\textit{VoxSeT + CT3D (RoI head)} &Transformer& \textbf{70.21} &\textbf{92.05} &\textbf{70.10}&\textbf{43.20} &80.56 &96.79&80.44& 62.37
				
				\\ \hline

			\end{tabular}
		\end{adjustbox}
		\label{tbl:waymo_val}
	
	\end{table*}

	\begin{table*}[t]
		\caption{Performance comparison with traditional single-stage baseline models on the KITTI validation set. The results are reported by the mAP with 11 recall points.}
			\vspace{-2mm}
		\centering
		
		\begin{adjustbox}{width=0.96\textwidth}
			\begin{tabular}{c||ccc||ccc||ccc}
				\hline
				\multirow{2}{*}{Method} &
				\multicolumn{3}{c||}{Vehicle } &
				\multicolumn{3}{c||}{Pedestrian }  &
				\multicolumn{3}{c}{Cyclist }  \\
				
				\cline{2-10}
				
				{}  & Easy & Moderate  & Hard & Easy & Moderate  & Hard  & Easy & Moderate  & Hard  \\ \hline\hline
				SECOND \cite{second} & 88.61 & 78.62 & 77.22 & 56.55  & 52.98 &47.73 &80.58 &67.15 & 63.10 \\
				PointPillars \cite{pointpillars} & 86.46 & 77.28 & 74.65 & 57.75 & 52.29 & 47.90 & 80.04 & 62.61 &59.52 \\
				VoxSeT (single-stage) & 88.45&78.48&77.07 &60.62 &54.74&50.39&84.07&68.11&65.14 \\ \hline
				Improvements & -0.16&-0.14&-0.15 &2.87 &1.76&2.49&3.49&0.96&2.04 \\
				\hline	
				
		\end{tabular}
	\end{adjustbox}
	\label{tbl:kitti}
	\vspace{-2mm}
	\end{table*}

	\begin{table}[t]
		\caption{Performance comparison with state-of-the-art methods on the KITTI test set. The results are reported by the mAP with 0.7 IoU threshold and 40 recall points.}
		\vspace{-2mm}
		\centering
		\begin{adjustbox}{width=0.99\columnwidth}
			\begin{tabular}{c|ccc}
				\hline
				\multirow{2}{*}{Method} &
				
				\multicolumn{3}{c}{\textit{3D}} \\
				
				\cline{2-4}
				
				{} & Easy & Moderate  & Hard   \\ \hline\hline
				\textit{\textbf{LiDAR + RGB}}: & & & \\
				MV3D\cite{mv3d} (CVPR17)  &74.97 & 63.63 & 54.00 \\
				ContFuse\cite{contfuse} (ECCV18)  &83.68 &68.78& 61.67 \\
				AVOD-FPN\cite{avod} (IROS18) &83.07 &71.76 & 65.73 \\
				F-PointNet\cite{frustum} (CVPR18) &82.19& 69.79 &60.59 \\
				MMF \cite{mmf} (CVPR19) &88.40&77.43 &70.22 \\
				3D-CVF\cite{3d-cvf} (ECCV20)  &89.20 & 80.05 & 73.11 \\
				CLOCs \cite{clocs} (IROS20) &88.94 &80.67 &77.15 \\ \hline\hline
				\textit{\textbf{LiDAR only}}: & & & \\
				VoxelNet\cite{voxelnet} (CVPR18) &77.47 &65.11 &57.73 \\
				SECOND\cite{second} (Sensor18) &83.34 & 72.55 &65.82 \\
				PointPillars\cite{pointpillars} (CVPR19) & 82.58 &74.31 &68.99 \\
				STD\cite{std} (ICCV19) & 87.95 &79.71& 75.09 \\
				
				PointRCNN\cite{pointrcnn} (CVPR19)  &86.96 & 75.64 &70.70 \\
				SA-SSD\cite{sa-ssd} (CVPR20)  & 88.75 &79.79 &74.16 \\
				3DSSD\cite{std} (CVPR20)  & 88.36 &79.57 &74.55\\
				PV-RCNN\cite{std} (CVPR20)  & \textbf{90.25} &81.43 &76.82 \\
				Voxel-RCNN\cite{std} (AAAI21)  & 87.95 &79.71& 75.09 \\
				CT3D \cite{ct3d} (ICCV21) &	87.83 & 81.77 &	77.16 \\
				VoTR-TSD \cite{votr} (ICCV21) & 89.90 &	\textbf{82.09} & \textbf{79.14} \\
				\hline \hline
				VoxSeT (ours) &  88.53 &82.06 &	77.46  \\ \hline
				
			\end{tabular}
		\end{adjustbox}
		\label{tbl:kitti_test}
		\vspace{-3mm}
	\end{table}

	\section{Experiments}
	
	In this section, we evaluate our proposed VoxSeT on two public detection datasets, KITTI \cite{kitti} and Waymo \cite{waymo}. We first introduce the training details of VoxSeT and the evaluation settings, and then compare our models with state-of-the-art detection models. Finally, we conduct an in-depth analysis of each component of VoxSeT.
	
	\subsection{Implementation details} \label{sec:settings}
	
	%\noindent 
	\textbf{Model setup.} On the KITTI dataset, we select the LiDAR points that fall into the ranges [0m, 70.4m], [-40m, 40m], [-3m, 1m] along X, Y, Z axes, respectively, and abandon those points with the frontal view projections out of image. On the Waymo dataset, the points that lie between [-75.2m, 75m] in the X and Y axes, and [-2m, 4m] in the Z axis are selected. The voxel size of the first VSA layer is [0.32m, 0.32m, 4m] on KITTI and [0.32m, 0.32m, 6m] on Waymo. The voxel size is doubled along the X and Y axes in the next VSA block. The feature dimensions of the four VSA blocks are 16, 32, 64 and 128, respectively. The number of latent codes in each VSA block is 8 and the bandwidth $L$ of the Positional Embedding (PE) module is 64. 
	
	%\noindent 
	\textbf{Training and inference.}
	The network is trained end-to-end on four RTX Quodra 8000 GPUs for 100 epochs with the Adam optimizer. The batch size, learning rate, and weight decay are set to 4, 0.003 and 0.01, respectively. The learning rate is decayed with the $onecycle$ policy, where the momentum has a damping range of $[85\%, 95\%]$. 
	
	We apply the anchor settings in SECOND \cite{second} in our single stage model. For the two-stage model, we sample 512 RoIs in training and 128 RoIs in inference. In the post-processing phase, the bounding-boxes are filtered by NMS with an IoU threshold of 0.1, and those having confidence over 0.3 are selected as final predictions. Data augmentations \cite{voxelnet,second,fast-pointrcnn} are applied to improve the model generalization performance. For other default settings, the readers are referred to the OpenPCDet toolbox \cite{openpcdet} used in this work.
	
	\subsection{Dataset and evaluation metrics}
	%\noindent 
	\textbf{KITTI dataset \cite{kitti}.} KITTI contains 7,481 training samples and 7,518 testing samples. Following the common protocol \cite{mv3d}, we split the labeled data into a training set with 3,712 samples and a validation set with 3,769 samples. We conduct experiments on the commonly used car category whose detection IoU threshold is 0.7, and report the results on three difficulty levels (\textit{easy}, \textit{moderate} and \textit{hard}) according to the object size, occlusion state and truncation level. 
	
	%\noindent 
	\textbf{Waymo open dataset \cite{waymo}.} 
	This dataset consists of 798 training sequences and 202 validation sequences, where there are 158,361 samples and 40,077 samples, respectively. The evaluation metrics used are 3D mean Average Precision (mAP) with IoU threshold of 0.7 on the vehicle category. The measures are reported based on the distances from objects to sensor, \ie, 0$-$30m, 30$-$50m and $>$50m, respectively. Two difficulty levels, LEVEL 1 (boxes with more than five LiDAR points) and LEVEL 2 (boxes with at least one LiDAR point) are considered.
	
	%-------------------------------------------------------------------------

	\subsection{Results on the Waymo open dataset}
	We first evaluate the performance of VoxSeT on the Waymo open dataset. The results are summarized in Table \ref{tbl:waymo_val}. VoxSeT outperforms most of CNN-based models, leading PV-RCNN \cite{pvrcnn} by 5\% LEVEL\_1 mAP and VoxelRCNN \cite{voxel-rcnn} by 2.4\% LEVEL\_2 mAP. As one of the few transformer based models, our VoxSeT achieves better performance than its transformer-based competitor VoTR-TSD\cite{votr}, which brings 0.9\% and 1.4\% improvements on LEVEL\_1 and LEVEL\_2 mAP, respectively. VoxSeT achieves comparable performance to the state-of-the-art method CT3D \cite{ct3d}. It should be noted that, however, CT3D actually employs a heavy transformer based RoI head, which has three self-attention encoding layers. By adopting this RoI head into VoxSeT, our model achieves better results, outperforming CT3D by 1.5 \% LEVEL\_1 mAP and 1.2\% LEVEL\_2 mAP. This demonstrates that as a new transformer based backbone network, VoxSeT surpasses Sparse CNN based networks. VoxSeT works especially well in the range of 30-50m, which indicates that transformer modules are better in capturing the context information in long-range areas. 
	
	\begin{table}[h]
		\caption{Performance comparison with state-of-the-art methods on the KITTI validation set. The results are reported by the mAP with 0.7 IoU threshold and 11 recall points.}
		\vspace{-2mm}
		\centering
		\begin{adjustbox}{width=0.99\columnwidth}
			\begin{tabular}{c|ccc}
				\hline
				\multirow{2}{*}{Method} &
				
				\multicolumn{3}{c}{\textit{3D}} \\
				
				\cline{2-4}
				
				{}  & Easy & Moderate  & Hard   \\ \hline\hline
				\textit{\textbf{LiDAR + RGB}}: & & & \\
				MV3D\cite{mv3d} (CVPR17)   &71.29 &62.68 &56.56 \\
				
				F-PointNet\cite{frustum} (CVPR18)   &83.76 & 70.92 &63.65 \\
				
				3D-CVF\cite{3d-cvf} (ECCV20)   &89.67 &79.88 &78.47 \\ \hline\hline
				
				\textit{\textbf{LiDAR only}}: & & & \\
				SECOND\cite{second} (Sensor18) &88.61 &78.62 &77.22\\
				PointPillars\cite{pointpillars} (CVPR19)  &  86.62 &76.06 &68.91 \\
				STD\cite{std} (ICCV19)  &89.70 &79.80 &\textbf{79.30} \\
				
				PointRCNN\cite{pointrcnn} (CVPR19)  &88.88& 78.63 &77.38 \\
				SA-SSD\cite{sa-ssd} (CVPR20)  & \textbf{90.15} &79.91 &78.78 \\
				3DSSD\cite{std} (CVPR20)  & 89.71 &79.45 & 78.67\\
				PV-RCNN\cite{std} (CVPR20) & 89.35 &83.69 &78.70 \\
				Voxel-RCNN\cite{std} (AAAI21)  & 89.41 &84.52 &78.93\\
				CT3D \cite{ct3d} (ICCV21) &	89.54 &86.06 &78.99\\
				VoTR-TSD \cite{votr} (ICCV21)  & 89.04 &84.04& 78.68\\
				\hline \hline
				VoxSeT (ours) & 89.21 &\textbf{86.71} &	78.56  \\ \hline
				
			\end{tabular}
		\end{adjustbox}
		\label{tbl:kitti_val}
		\vspace{-3mm}
	\end{table}

	\subsection{Results on the KTTI Dataset}
	We then conduct experiments on the KITTI dataset to evaluate the performance of VoxSeT as a single-stage detection model. Our competitors are SECOND \cite{second} and PointPillars \cite{pointpillars}, which represent two widely used baseline feature extractors. All the three methods use the same detection head and hyper-parameters in training. As shown in Table \ref{tbl:kitti}, VoxSeT achieves comparable performance to SECOND on vehicle class, but much better performance on Pedestrian and Cyclist classes. We believe this is because VoxSeT has a wider effective receptive field through the VSA conditioned on the large voxel, which is crucial to detecting the objects with sparse points. 
	
	As a two-stage detection model, VoxSeT achieves better performance than CT3D by 0.7\% (Easy), 0.4\% (Moderate) and 0.3\% (Hard) mAP, respectively, as shown in Table \ref{tbl:kitti_test}. Compared with VoTR-TSD which relies on multi-scale backbone features, VoxSeT can still achieve comparable performance by using only singular point-wise features. We also evaluate VoxSeT on KITTI $val$. One can see that VoxSeT achieves leading accuracy on ``Moderate'' level but slightly lower accuracy on ``Easy'' and ``Hard'' levels. We believe this is because KITTI has a long-tailed distribution, and hence the ``Moderate'' samples dominate the hidden space of the VSA module.
	
	It should be pointed out that both CT3D and VoTR employ convolutional architectures, and their performances on KITTI and Waymo datasets are not consistent. Specifically, CT3D works better on Waymo but its performance drops much on KITTI, while VoTR works better on KITTI but its performance on Waymo is much worse. In contrast, our VoxSeT exhibits consistently superior performance on both datasets, demonstrating its good generalization capacity.
	
%	\begin{table}
%		\caption{The mAP (11 recall points) on KITTI $val$ by using constant voxel size. The mAP is based on the 3D bounding-boxes produced by RPN. The default setting uses incremental voxel size.}
%		\vspace{-3mm}
%		\centering
%		%\begin{adjustbox}{width=0.95\columnwidth}
%		\begin{tabular}{c|cccc|c}
%			\hline
%			voxel size          & 0.2m  & 0.4m & 0.8m & 1.6m & default\\  \hline
%			mAP            & 49.2 & 66.29  & 68.17 & 69.41 &   79.56 \\ \hline 
%		\end{tabular}
%		%\end{adjustbox}
%		\label{tbl:voxel-size}
%	    \vspace{-1mm}
%	\end{table}
	
	\begin{table}
		\caption{Effects of enabling token interactions by convolutional feed-forward network. The mAP (11 recall points) of RPN on KITTI $val$ are reported.}
		\vspace{-2mm}
		\centering
		\begin{adjustbox}{width=0.8\columnwidth}
			\begin{tabular}{c||ccc}
				\hline
				Settings          & Easy  & Moderate & Hard \\  \hline
				
				default & 88.31 & 79.56  & 77.84  \\ 
				ConvFFN$\rightarrow$FFN            & 70.12 & 69.54  & 54.23  \\ \hline
				
			\end{tabular}
		\end{adjustbox}
		\label{tbl:conv-ffn}
	
	\end{table}
	
	\begin{table}
		\caption{The mAP (11 recall points) of RPN on KITTI $val$ by using different number of latent codes (LC) in four VSA modules.}
		\vspace{-2mm}
		\centering
		%\begin{adjustbox}{width=0.95\columnwidth}
		\begin{tabular}{c|ccc}
			\hline
			No. of latent codes   & 4 & 8 & 16\\  \hline
			mAP           & 76.63  & 78.25 &  78.74 \\ \hline 
		\end{tabular}
		%\end{adjustbox}
		\label{tbl:latent}
		
	\end{table}

	\begin{table}
		\caption{Comparison of PointRCNN \cite{pointrcnn} detectors with VoxSeT and PointNet++ backbones. The mAP (11 recall points) on KITTI $val$ are reported.}
		\vspace{-2mm}
		\centering
		\begin{adjustbox}{width=0.8\columnwidth}
			\begin{tabular}{c||ccc}
				\hline
				Settings          & Easy  & Moderate & Hard \\  \hline
				
				\textit{PointRCNN} &88.52 & 78.95 & 77.81 \\ 
				VSA-\textit{PointRCNN}            & 89.61 & 80.14  & 78.69 \\ \hline
				
			\end{tabular}
		\end{adjustbox}
		\label{tbl:pointrcnn}
		
	\end{table}
	
	\begin{table}
		\caption{The latency and runtime memory on the KITTI dataset, tested by NVIDIA 2080Ti GPU. }
		\vspace{-2mm}
		\centering
		\begin{adjustbox}{width=0.9\columnwidth}
			\begin{tabular}{c||cc}
				\hline
				Models          & Latency  & Memory (runtime) \\  \hline
				
				SECOND \cite{second} & 48 ms & 6093MB    \\ 
				PointPillars \cite{pointpillars} &  22ms & 1508 MB \\
				VoxSeT (single-stage) & 34 ms & 2381 MB \\ \hline
				
			\end{tabular}
		\end{adjustbox}
		\label{tbl:runtime}
		\vspace{-1mm}
	\end{table}

	\begin{figure*}[ht!]
		\centering
		\includegraphics[width=0.01\textwidth]{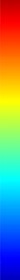}
		\includegraphics[width=0.24\textwidth]{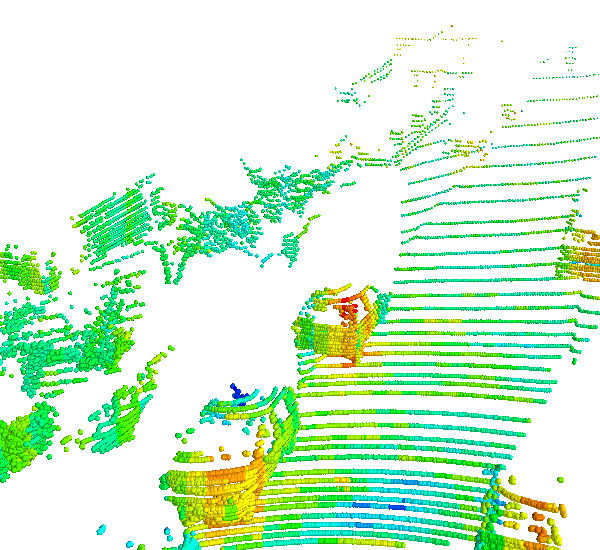}	
		\includegraphics[width=0.24\textwidth]{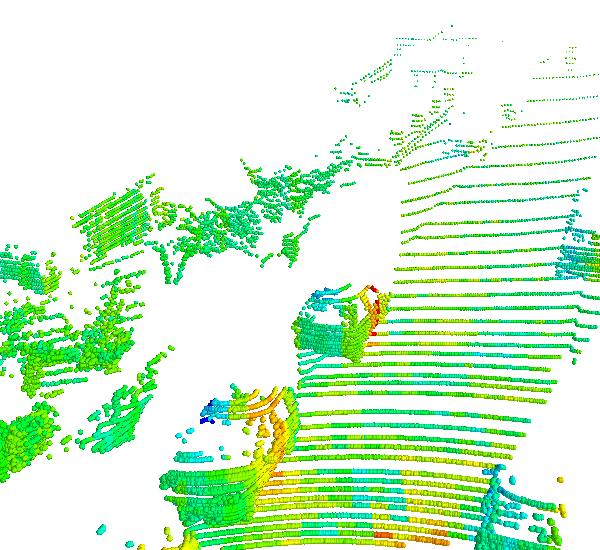}
		\includegraphics[width=0.24\textwidth]{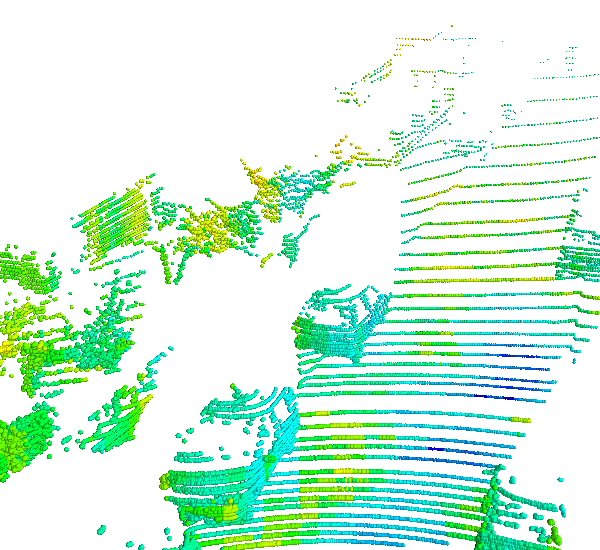}
		\includegraphics[width=0.24\textwidth]{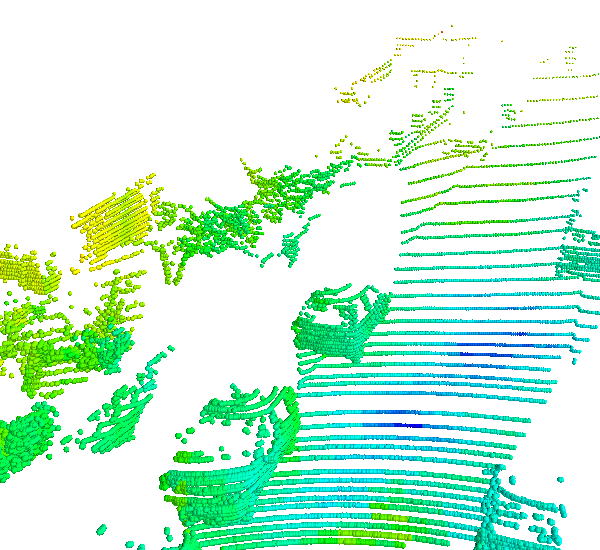}
		\vspace{-2mm}
		\caption{Visualization of the spatial attention maps induced by different latent codes in the last VSA module.}
		\vspace{-2mm}
		\label{fig:attn}
	\end{figure*}

	\subsection{Ablation study}
	\label{sec:ablation}
	We conduct a series of ablation experiments to comprehend the roles of different components in VoxSeT.
	
%	\textbf {Incremental voxel size.} 
%	VoxSeT learns point-wise features in a set-to-set manner, thus it does not rely on any down-sampled layers to increase the semantic levels. VoxSeT encodes different levels of context by gradually increasing the voxel size. To justify this, we test the performance by using constant voxel sizes and use the RPN output to calculate the mAP results. As shown in Table \ref{tbl:voxel-size}, though larger voxel sizes can capture richer context information, using constant voxel size in the backbone will degrade the performance by 10 mAP. This shows that gradually increasing the context level is a key to reliable feature extraction.
	
	\textbf{Convolutional feed-forward network.} 
	Table \ref{tbl:conv-ffn} shows that replacing the proposed ConvFFN with the conventional FFN significantly degrades the accuracy, indicating that the local connectivity is crucial to the detection performance.
	
	\textbf{Effects of number of latent codes.}
	In Table \ref{tbl:latent}, we investigate the number of latent codes used in four VSA modules. We see that more latent codes can encode more context information of point cloud, and enhance the modeling capacity of VoxSeT. 
	
	\textbf{Comparison with PointNet++ backbone.}
	We train a PointRCNN \cite{pointrcnn} variant by replacing its PointNet++ backbone \cite{pointrcnn} with our VoxSeT backbone. From Table \cite{pointnet++}, one can observe obvious performance improvements. We believe this is because VSA module has better modeling power in terms of dynamic learning and large receptive field than the set abstraction (SA) module in PointNet++.
	
	\textbf{Latency and runtime memory.}
	Table \ref{tbl:runtime} shows that VoxSeT is faster and has less memory consumption compared to the sparse 3D CNN (SECOND). The higher performance than PointPillars and the acceptable runtime cost suggest that VoxSeT can be a good alternative to PointPillars in real-time applications.

	\textbf{Visualization of attention weights.}
	Figure \ref{fig:attn} visualizes the spatial attention maps for the latent codes in the last VSA module. We show the attention maps for 4 out of a total of 8 latent codes. One can observe that the VSA module focuses more on the object region and different latent codes encode different contexts of the objects, indicating the high expressiveness of VSA for point cloud data.

	\section{Conclusion and discussions}
	We proposed VoxSeT, a novel transformer-based framework for 3D object detection
	from LiDAR point clouds. In contrast to previous 3D LiDAR detectors, which use sparse CNN and PointNet backbones to learn point cloud features, we made the first attempt to model point cloud processing as set-to-set translation, which preserves the full resolution of raw point cloud at every step of feature extraction. We presented a voxel-based set attention module that performs self-attention on voxel clusters of arbitrary size and encodes point features with more discriminative context information from a large receptive field. Experimental results on the Waymo and KITTI datasets demonstrated that our VoxSeT can achieve competitive performance, making it a good alternative for point cloud modeling. 
	
	It should be noted that in VoxSeT, we only explored one possible formulation of liner attention based on the induced latent codes. This limits the expressive power of VoxSeT to represent different point cloud structures and their correlations. By using stronger attention mechanisms, the performance of VoxSeT can be further improved, which will be our future research direction.

	%%%%%%%%% REFERENCES
	{\small
		\bibliographystyle{ieee_fullname}
		\bibliography{egbib}
	}
	
\end{document}